\documentclass[a4paper]{journal}
\usepackage[utf8]{inputenc}
\usepackage{amsmath}
\usepackage{authblk}

\newcommand{\pkg}[1]{{\normalfont\fontseries{b}\selectfont #1}}

\let\code=\texttt

\title{FREEtree: A Tree-based Approach for High Dimensional Longitudinal Data With Correlated Features}
\author[1]{\large Yuancheng Xu}
\author[2]{\large Athanasse Zafirov}
\author[3]{\large R. Michael Alvarez}
\author[4]{\large Dan Kojis}
\author[5]{\large Min Tan}
\author[6]{\large Christina M. Ramirez}

\affil[1]{Department of Mathematics, Southern University of Science and Technology}
\affil[2]{Anderson School of Management, University of California, Los Angeles}
\affil[3]{Division of Humanities and Social Sciences, California Institute of Technology}
\affil[4]{Department of Statistics, The University of Wisconsin–Madison}
\affil[5]{Department of Mathematics, Sichuan University}
\affil[6]{Department of Biostatistics, Fielding UCLA School of Public Health}

\date{}
\Authfont{\itshape\tiny}

\usepackage[numbers]{natbib}
\usepackage{graphicx}

\usepackage{bbm}
\usepackage{amsmath}
\usepackage{tikz} 
\usepackage{booktabs}
\usepackage{multirow}
\usepackage{siunitx}

\begin{document}

\maketitle

\begin{abstract}
This paper proposes FREEtree, a tree-based method for high dimensional longitudinal data with correlated features. Popular machine learning approaches, like Random Forests, commonly used for variable selection do not perform well when there are correlated features and do not account for data observed over time.  FREEtree deals with longitudinal data by using a piecewise random effects model. It also exploits the network structure of the features by first clustering them using weighted correlation network analysis, namely WGCNA. It then conducts a screening step within each cluster of features and a selection step among the surviving features, that provides a relatively unbiased way to select features. By using dominant principle components as regression variables at each leaf and the original features as splitting variables at splitting nodes, FREEtree maintains its interpretability and improves its computational efficiency. The simulation results show that FREEtree outperforms other tree-based methods in terms of prediction accuracy, feature selection accuracy, as well as the ability to recover the underlying structure. \let\thefootnote\relax\footnotetext{The R package is currently undergoing the CRAN submission process and will soon be freely available on the their repository. You can currently install it in R using  \texttt{install\char`_github("adzafirov/FREEtree", force = TRUE ) }or access it through GitHub via \texttt{https://github.com/adzafirov/FREETree}
 }
\end{abstract}

\section{Introduction}
Longitudinal or clustered data, where observations within a unit (cluster) are more correlated than observations from other units (clusters), are very common in areas such as social science and medical research. Further, the data may contain a large number of correlated features relative to the number of observations (high dimensional data). The goal of this paper is to extend tree-based algorithms to high dimensional longitudinal data with correlated features and to develop a relatively interpretable data mining technique for feature selection and prediction.

Tree based algorithms began to gain momentum with the appearance of  the CART (classification and regression trees) algorithm  (Breiman et al., 1984) \cite{breiman1984classification}.
They are widely used in statistical machine learning due to their interpretability, relatively high computational efficiency, and their nonparametric and nonlinear nature. Briefly, a binary decision tree-based algorithm recursively partitions the parameter space into relatively pure nodes using a splitting criterion such as entropy or Gini impurity by searching every possible variable, and each possible split point, until it meets a pre-specified stopping criteria, and builds a piece-wise model on each subset of the data.  The algorithm is greedy by nature and does not take into account correlation or longitudinal structure.

Segal \cite{segal1992tree} made the first attempt to deal with longitudinal data by using regression trees, and proposing a new split function depending on the covariance structure of multiple responses. However, this method cannot deal with time-varying covariates (only the responses, and not the covariates, vary with time in his setting) and all the observations within a unit end up in one terminal node. Mixed-effects longitudinal trees (MELT)(Cho et al., 2014) \cite{eo2014tree} fully explore the shape of the data with respect to time by fitting low degree polynomials and splits on the coefficients. The objective of MELT is to identify different shapes of time among units. However, MELT only deals with time-invariant covariates, and is not optimized for prediction. 

Sela and Simonoff (2012) \cite{sela2012re} proposed the RE-EM tree, which uses a random effects model to deal with longitudinal structure, where the fixed effect is modelled as a standard regression tree CART. The random effects and fixed effects are estimated alternatively, which is similar to the EM algorithm. Later, a new version of RE-EM tree was proposed by Simonoff and Fu (2015) \cite{fu2015unbiased} where the implementation of the fixed effect was replaced by the conditional inference trees of Hothorn et al. (2006) \cite{hothorn2006unbiased} to reduce bias. RE-EM tree can deal with time-varying covariates, and observations within a unit can end up in different terminal nodes. 

The generalized
linear mixed-effects model trees (GLMM tree) algorithm (M. Fokkema et al., 2017) \cite{fokkema2018detecting} adopts a more general approach than the RE-EM tree. The GLMM tree also uses a random effects approach, but with the fixed effect modelled as a piece-wise generalized linear model, that is, as a regression model tree with a generalized linear model, instead of a constant, at each leaf. The fixed effects and random effects are estimated in an alternative way, with one estimated after the other until
convergence.  The GLMM tree provides more flexibility in the model of fixed effect and can be used to detect treatment effects (see \ref{sec:design of simulations}).  The GLMM tree approach will be discussed in more detail in \ref{sec:GLMM Tree}.

It is known that Random Forest variable selection is biased when there is correlation among the features. Fuzzy Forests (Conn, Ramirez et al., 2015) \cite{conn2015fuzzy} was developed to address correlation within the predictors in the setting where the number of parameters is much greater than the number of observations ($p >> n$).  The first step in Fuzzy Forests is to explicitly cluster features using weighted correlation networks \cite{zhang2005general} (reviewed in \ref{sec: WGCNA}). Then a feature screening step is conducted within each cluster using Recursive Feature Elimination Random Forests (RFE-RFs) \cite{diaz2006gene}. Finally a feature selection step is done within the features selected from the screening step, allowing clusters to interact with each other. The screening step and the selecting step enables Fuzzy Forests to select features in a relatively unbiased way in the presence of highly correlated features.  The Fuzzy Forests methodology has been used in a number of applied research articles, for example \cite{conn_ramirez2016, kim-alvarez-ramirez, ramirez-abrajano-alvarez2019}.

This paper proposes the Fuzzy Random Effect Estimation tree (FREEtree), which takes advantage of the powerful feature selection approach of Fuzzy Forests, as well as the flexible framework of the GLMM tree, to deal with the longitudinal structure of data. 

The remainder of the article is organized as follows: Section \ref{sec: review of WGCNA and GLMM} reviews the building blocks of FREEtree before section \ref{sec: FREEtree method} explains the FREEtree algorithms in detail. Section \ref{sec: Simulation results} provides simulation results of FREEtree on two simulated data sets, one with a time-treatment interaction and one without. Section \ref{Discussion and future work} discusses future research development of FREEtree and the last section concludes the paper.

\section{A review of WGCNA and GLMM tree}
\label{sec: review of WGCNA and GLMM}
\subsection{WGCNA}
\label{sec: WGCNA}
Weighted correlation networks (WGCNA) have been used in many applications to examine the network structure of covariates \citep{langerfelder2008wgcna, pei2017wgcna,Gudenas2015wgcna}.  This is an unsupervised learning method.  
In order to construct the network, WGCNA does the following: (1) choose a similarity function for feature $X^{u}$ and $X^{v}$, denoted by $s_{uv}$. A common choice is $Corr(X^{u},X^{v})$ where Corr is the Pearson correlation. Then compute the similarity matrix $S = [s_{uv}]$. (2) Transform the similarity matrix X by the adjacency matrix $A = [a_{uv}]$ where $a_{uv} = s_{uv}^{\beta}$ which results in a soft-thresholding network. The $\beta$ is chosen according to the scale-free criterion \citep{zhang2005general}.(3) Convert the adjacency matrix A to the topological overlap matrix (TOM) W through Eq.\eqref{eq:TOM} where $q_{uv}=\sum_{r=1}^{p}a_{ur}{a_{rv}}$ and $c_{u} = \sum_{r=1}^{p} a_{ur}$. (4) Use a hierarchical clustering tree algorithm to find clusters using TOM. The reason that hierarchical clustering algorithm uses TOM instead of the adjacency matrix $A$ is that using TOM may lead to more distinct modules \citep{zhang2005general}.
\begin{equation}\label{eq:TOM}
    w_{uv} = \frac{q_{uv}+a_{uv}}{\min\{c_{u},c_{v}\}+1-a_{uv}}
\end{equation}

Weighted correlation network analysis (WGCNA)\citep{zhang2005general} can be used for clustering covariates where covariates within each module are highly correlated and features (or g, we will often use the term genes in this paper to remain consistent with the genetics literature) from different modules are approximately uncorrelated. Covariates that are not assigned to any clusters are placed in the grey module. That is, each grey covariate in the grey module is roughly uncorrelated to any other covariates and can be viewed as a cluster on its own. Thus, note that in the context of machine learning, we can view each feature as a gene and therefore WGCNA can identify modules of highly correlated features.

\subsection{GLMM tree}
\label{sec:GLMM Tree}
The rational behind the Generalized Linear Mixed-Effects Model tree (GLMM tree) \citep{fokkema2018detecting} is that a global generalized linear mixed-effect model may not fit the data well. However, if additional splitting variables are available, we can fit the data with piece-wise models by partitioning the data with these splitting variables.

For example, suppose that in our dataset the $t$\textsuperscript{th} observation of cluster $i$ consists of covariates $x_{it}$ and response $y_{it}$. Cluster $i$ may stand for the i\textsuperscript{th} patient and $t$, the time of the measurement. Then a global Generalized Linear Mixed-Effects model (GLMM) is given by
\begin{equation}\label{eq:global GLM}
  E[y_{it}|x_{it}] = \mu_{it};\quad g(\mu_{it}) = x_{it}^{T}\beta + z_{i}^{T}b_{i};
\end{equation}
where $g$ is the link function and $\beta$ is a vector of fixed-effect regression coefficients (as opposed to the power function described in WGCNA). For a mixed-effect model with only a random intercept, $z_{i}$ is just constant (1) and $b_{i}$ is the random intercept associated with cluster $i$. When random slopes are involved, $z_{i}$ is the design vector which is a subset of $x_{it}$ and $b_{i}$ is the random vector with each component corresponding to the random deviation of the slope from the fixed-effect. For simplicity, we assume that the link function $g$ is the identity function and the mixed-effects model with only random intercept is adopted. That is, we are using a linear mixed-effect model with only a random intercept from now on, as the following:
\begin{equation}\label{eq:global LMM}
 \mu_{it} = x_{it}^{T}\beta + b_{i}
\end{equation}

In many cases, the Linear Mixed-Effect model (LMM) in Eq\eqref{eq:global LMM} may not fit the data well because the assumption that the underlying fixed-effect model is a linear function is too restrictive. It often makes more sense to approximate the fixed-effect structure with a piece-wise linear model instead of a global linear model.  GLMM tree uses a model-based recursive partitioning (MOB) algorithm \citep{zeileis2008model} that partitions the dataset using splitting variables and find better-fitting local LMM models. MOB iterates the following: fit a parametric model (such as LMM) to the dataset and then adopt parameter stability tests on each of splitting variables by computing a p-value for every splitting variable. If the smallest p-value is below the significant level $\alpha$, the dataset is split into two subsets using the splitting variable value with the smallest p-value, with the split point for that variable chosen to minimize the instability. Therefore only significant splitting variables will be used for splitting at the node of a GLMM tree. More details of the parameter stability test are described by Zeileis \cite{zeileis2008model}. The resulting GLMM tree has the following form, 

\begin{equation}\label{eq:local LMM}
 \mu_{it} = x_{it}^{T}\beta_{j(it)} + b_{i}
\end{equation}
where $j(it)$ is the index of terminal node that $t$\textsuperscript{th} observation of cluster $i$ belongs to. Note that the fixed-effect is now a piece-wise linear function of covariates and the random intercept is global in the sense that it only depends on the cluster, instead of the terminal node. The GLMM tree is trained by iteratively estimating the fixed-effect (a linear mixed-effects tree) assuming random effects are known and estimating random effects by assuming fixed effects are known until convergence.

The R package, \pkg{glmertree} \citep{fokkema2018detecting} implements the GLMM tree.  In the following section, we use LMM tree for simplicity. That is, we assume the link function $g$ is the identity function. The function \code{lmertree()} is used in this package.

\section{The FREEtree estimation method}
\label{sec: FREEtree method}
The goal of FREEtree is feature selection and then using selected features to make predictions.  The advantage of having fewer features is parsimony and increased interpretability.  At the heart of the algorithm lies a binary decision tree splitting strategy that is easily interpretable. While CART and many other methods are usually biased towards selecting correlated features while ignoring independent ones in feature selection, FREEtree reduces this bias by clustering features by their correlation pattern and screening features within each cluster, while allowing for features to interact. The resulting features are used to fit a LMM tree, which includes a linear regression model at the end of each leaf that also considers a random effect at the patient level. The predictive power mostly comes from LMM tree, which fits the data with a piecewise linear function of covariates plus a random effect, instead of a piece-wise constant function like CART and RE-EM tree. However, in order to regress on covariates, feature selection is necessary because linear regression requires that the sample size be sufficiently larger than the number of parameters for identifiability. FREEtree integrates feature selection and prediction in a natural way and is particularly useful when $p$ is larger than $n$.

\subsection{Notation}
The training dataset consists of patients $i = 1,2,...,n$, who are measured at time $t = 1,2,..,T$. To simplify the notation, we assume balanced data here, though this is not required for the FREEtree algorithm. Each patient has three types of features:
\begin{itemize}
    \item \texttt{var\char`_select} X: Features of length $p$ that will be chosen from.
    \item \texttt{fixed\char`_regress} R: Features that will be used for regression in every tree. In longitudinal settings, this could be time or higher order of time.
    \item \texttt{fixed\char`_split} S: Features that will be considered as splitting variables in every tree.
\end{itemize}
The value of features of patient $i$ at time $t$ is denoted by $x_{it}$, $r_{it}$ and $s_{it}$ respectively. Note that \texttt{var\char`_select}, \texttt{fixed\char`_regress} and \texttt{fixed\char`_split} can be empty. The selection of which type each features belongs in is left up to the user. The goal of FREEtree is to select important features from \texttt{var\char`_select} and use the selected features as well as \texttt{fixed\char`_regress} and \texttt{fixed\char`_split} to give the final prediction.

\subsection{The FREEtree algorithm}
\label{sec:Fuzzy Strategy}
The FREEtree algorithm consists of a feature selection step and a prediction step. First assume that \texttt{fixed\char`_regress} is not empty. The case where it is empty will be discussed in section \ref{sec:PC Strategy}. 

The feature selection step has three steps: clustering, screening, and selection. During the clustering step, features in \texttt{var\char`_select} are clustered by WGCNA into modules, which includes a grey module and non-grey modules whose number is not known a priori. The grey module includes all covariates that have low connectivity and can be viewed as roughly independent.  Features within the same non-grey module are highly correlated/connected with each other and have lower correlation or connectivity with the features from other modules. Let there be $m$ modules selected by WGCNA.  Denote the modules of \texttt{var\char`_select} by $\{P_{1},...,P_{m}\}$ and let $p_{l} = |P_{l}|$ so that $\sum_{l=1}^{m} p_{l} = p$. Without loss of generality, denote the last module $P_{m}$ as the grey module.

For the screening step, features are selected within each module as follows: For module $l$ ($l=1,2,...,m$), use \texttt{fixed\char`_regress} as regression variables and use $P_{l}$ as well as consider \texttt{fixed\char`_split} for splitting variables to fit a LMM tree. The selected features from module $l$ are the set of features $P_{l}^{S}$ used in the LMM tree that are not included in \texttt{fixed\char`_split}. The result of the screening step is a set of screened features $\{P_{1}^{S},...,P_{m}^{S}\}$. 

The final selection step allows the selected features from each modules to interact with each other. FREEtree uses all of the screened features $\{P_{1}^{S},...,P_{m}^{S}\}$ from the screening  step and treats \texttt{fixed\char`_split} as splitting variables, then uses \texttt{fixed\char`_regress} as regression variables to fit a LMM tree. The final selected features from \texttt{var\char`_select} are the features used by this LMM tree that are not included in \texttt{fixed\char`_split}, denoted by $x^{S}$.

Finally, at the prediction step, a LMM tree is fitted using \texttt{fixed\char`_split} and $X^{s}$ as splitting variables and using \texttt{fixed\char`_regress} and $X^{s}$ as regression variables. The prediction is provided by this final LMM tree. Note that the final selected features $X^{s}$ from \texttt{var\char`_select} are used both as splitting and regression variables, which fits the data in a more flexible way than just regressing on \texttt{fixed\char`_regress}.

\subsection{Another strategy for feature selection}
\label{sec:Non Fuzzy Strategy}

The screening and selection steps help reduce bias in feature selection by eliminating features in correlated modules and thus protecting independent features from being ignored by LMM tree. However, if the number of non-grey modules is large and there are many correlated features after screening step, the independent features are still in the danger of being ignored at the selection step. In order to help protect independent features, another strategy of feature selection is proposed, which is particularly helpful if the number of correlated feature is large compared with independent features. Users can set Fuzzy=False to use this strategy. If Fuzzy=True, the strategy in section \ref{sec:Fuzzy Strategy} will be adopted.

At the screening step, features within each non-grey modules $\{P_{1},...,P_{m-1}\}$ are screened into $\{P_{1}^{S},...,P_{m-1}^{S}\}$. That is, use $P_{l}$ ($l = 1,2,...,m-1$) and \texttt{fixed\char`_split} as splitting variables and use \texttt{fixed\char`_regress} as regression variables to fit a LMM tree and choose features $P_{l}^{S}$ used by the tree and not contained in \texttt{fixed\char`_split}. Note that for now we don't screen within the grey module $P_{m}$. Then we select features from within the set of screened features $\{P_{1}^{S},...,P_{m-1}^{S}\}$ from the non-grey groups by using all of the screened features and \texttt{fixed\char`_split} as splitting variables, and \texttt{fixed\char`_regress} as regression variables to fit a LMM tree. The selection step allows the non-grey modules to interact with each other producing  $\{Q_{1}^{S},...,Q_{m-1}^{S}\}$ with $Q_{l}^{S} \subset P_{l}^{S}$ for $l=1,2,...,m-1$. Then we fit a LMM tree using \texttt{fixed\char`_split} and features in the grey module as splitting variables, and regress on \texttt{fixed\char`_regress} as well as $\{Q_{l}^{S}\}_{l=1}^{m-1}$. The set of selected features from the grey module are the ones used in this LMM tree that are not included in \texttt{fixed\char`_split}, which is denoted by $Q_{m}^{S}$. The final result of feature selection is $\{Q_{l}^{S}\}_{l=1}^{m}$, denoted by $X^{s}$. A final LMM tree for prediction is fitted using \texttt{fixed\char`_split} and $X^{s}$ as splitting variables and using \texttt{fixed\char`_regress} and $X^{s}$ as regression variables.

\subsection{Use principal components in the absence of regressors}
\label{sec:PC Strategy}

Suppose that we do not have a natural choice for \texttt{fixed\char`_regress} and set it to empty. One obvious way to do feature selection and prediction is to use RE-EM tree \citep{sela2012re} with an averaged value at each leaf instead of a linear regression model. The disadvantage is that the assumption of the underlying true model being a RE-EM tree, a piece-wise constant function plus random intercept, can be too restrictive. 

It is more flexible to fit the underlying model with a piece-wise linear function in addition to a random intercept. Therefore another method that is proposed, which could have more power in feature selection and prediction, is to use the dominant principal components (PC) of the non-grey modules as intermediate regressors. The idea here is that in linear regression, using the dominant principle components as regressors has a comparable power in terms of prediction as using all the covariates as regressors, although interpretability is lost. However, FREEtree, even if it uses PCs, is still interpretable because PCs are used only in the step of feature selection and the selected features are determined by the non-terminal nodes of the tree, instead of PCs or any other regressors. The first PCs of non-grey modules are used for simplicity, though more dominant features can be used. Note that we do not use PCs of grey module since features within grey module are roughly independent and thus it is likely that there may be no dominant PCs.

For the screening step, features from non-grey modules $P_{l}$ (l=1,2,..,m-1) are selected by fitting a LMM tree using the first PC of $P_{l}$ as regression variables and use $P_{l}$ and \texttt{fixed\char`_split} as splitting variables. If Fuzzy=True, for the grey module $P_{m}$, a RE-EM tree is fitted using \texttt{fixed\char`_split} and the features used in the node of RE-EM tree are selected. Denote the screened features by $\{P_{l}^{S}\}_{l=1}^{m}$. For the selection step, final features $X^{S}$ are obtained by selecting from the screened features. That is, fit a RE-EM tree using $\{P_{l}^{S}\}_{l=1}^{m}$ and select those appeared in the nodes of the RE-EM tree. In the prediction step, a LMM tree is fitted using $X^{S}$ and \texttt{fixed\char`_split} as splitting variables and $X^{S}$ as regression variables.

If Fuzzy=False, final non-grey features $\{Q_{l}^{S}\}_{l=1}^{m-1}$ are obtained by selecting from screened features $\{P_{l}^{S}\}_{l=1}^{m-1}$ from non-grey modules. That is, use all the $\{P_{l}^{S}\}_{l=1}^{m-1}$ as splitting variables to fit a RE-EM tree and select features used in the node of RE-EM tree and not contained in \texttt{fixed\char`_split}. Then the selected grey-features $Q_{m}^{S}$ are obtained by fitting a LMM tree using the grey module $P_{m}$ and \texttt{fixed\char`_split} as splitting variables and $\{Q_{l}^{S}\}_{l=1}^{m-1}$ as regression variables. The final set of selected features $X^{S}$ is $\{Q_{l}^{S}\}_{l=1}^{m}$. The prediction is given by a LMM tree using $X^{S}$ and \texttt{fixed\char`_split} as splitting variables and using $X^{S}$ as regression variables.

\section{Simulation}
\label{sec: Simulation results}
\subsection{Design of simulations}
\label{sec:design of simulations}
We provide simulations to examine the utility of FREEtree in terms of feature selection, prediction and estimation of the underlying model structure. 
In all simulations, the training dataset has $n$ subjects (we will allow $n$ to vary) and each subject has $p=400$ features $X$ to be selected along with \texttt{fixed\char`_split} and \texttt{fixed\char`_regress}. The features, $X$, are grouped into 4 modules $\{X^{(1)},...,X^{(100)}\}$, \allowbreak  $\{X^{(101)},...,X^{(200)}\}$,$\{X^{(201)},...,X^{(300)}\}$ as well as $\{X^{(301)},...,X^{(400)}\}$. Each feature $X^{(i)}$ is generated from a multivariate normal distribution with mean 0 and variance 1. The features from different modules are uncorrelated and features within the first three modules are correlated with correlation 0.8, while features within the last module are uncorrelated. Therefore, the first three modules are called non-grey modules and the final module is the grey module, according to the conventions in WGCNA.

The first simulation includes a time by treatment interaction where different treatments corresponds to different patterns of response with respect to time. For simplicity, we assume two treatments here, $treatment_1$ and $treatment_2$. The true model for subject $i$ at time $t$ is given by $$y_{it}=f(X_{it})+(t-3)^2\mathbbm{1}_{treatment_1}-(t-3)^2\mathbbm{1}_{treatment_2}+b_{i}+\epsilon_{it}$$ where $\mathbbm{1}$ is the indicator function, $\epsilon_{it}$ is the error is drawn from normal distribution and $f$ is given by 
$$f(X) = 5X^{(1)}+2X^{(2)}+2X^{(3)}+5X^{(2)}X^{(3)}+5X^{(301)}+2X^{(302)}+2X^{(303)}+5X^{(302)}X^{(303)}$$
Here, only 6 variables out of 300 are important.  The other variables are noise.   We use $treatment$ as \texttt{fixed\char`_split} and use $time (t)$ and $time^2$ ($t^{2}$) as \texttt{fixed\char`_regress}. 
Here  \texttt{var\char`_select} is $X$ with important features being $X^{(1)},X^{(2)},\\X^{(3)},X^{(301)},X^{(302)}$ and $X^{(303)}$. Since we have a natural choice for \texttt{fixed\char`_regress}, in $time$ and $time^2$, we adopt the method described in
section \ref{sec:Fuzzy Strategy} and section \ref{sec:Non Fuzzy Strategy}.

In a second simulation, we consider a mixed effects model given by 
$$y_{it}=f(X_{it})+b_{i}+\epsilon_{it}$$ where $f$ and $\epsilon_{it}$ are the same as in the first simulation and $b_i$ is the random intercept corresponding to subject $i$ which is drawn from normal distribution with mean 0 and variance 3. Random intercepts of different subjects are independent. Since now we do not have a natural choice for \texttt{fixed\char`_regress}, we adopt the method described in section \ref{sec:PC Strategy}. That is, during the screening step, we regress on the first principal components of non-grey modules to select features from non-grey modules.

In both simulations, a validation set of 100 subjects is used for tuning parameters and a test set of 100 subjects is used for measuring root mean squared error on future observations. The prediction does not include random intercepts because they cannot be estimated from unknown patients. The performances of Random Forests and Fuzzy Forests in the following sections are measured by running the simulation 50 times using different random seeds.

\subsection{Predictive performance}
In this section, we first consider the dataset with the time-treatment interaction detailed in the previous section. We compare the predictive performance of FREEtree, Random Forests, Fuzzy Forests and LMM tree. For Random Forests and Fuzzy Forests, \texttt{var\char`_select} $\{X^{(v)}\}_{v=1}^{400}$,
\texttt{fixed\char`_regress} $time$ and $time^2$ and 
\texttt{fixed\char`_split} $treatment$ are used as covariates. $Time$, $time^2$ and $treatment$ are manually put into the "grey" module in Fuzzy Forests because the time variables are uncorrelated with $\{X^{(v)}\}_{v=1}^{400}$ in the generating process and treatment is categorical which WGCNA cannot deal with directly. For LMM tree, treatment and $\{X^{(v)}\}_{v=1}^{400}$ are specified as splitting variables and $time$, $time^2$ are used as the regression variables.
Note that unlike FREEtree,  we can not use all $\{X^{(v)}\}_{v=1}^{400}$  as regression variables because linear regression requires that the sample size be greater than the number of parameters in the linear regression model. 

\begin{figure}[htp]
    \centering
    \includegraphics[width=1\textwidth]{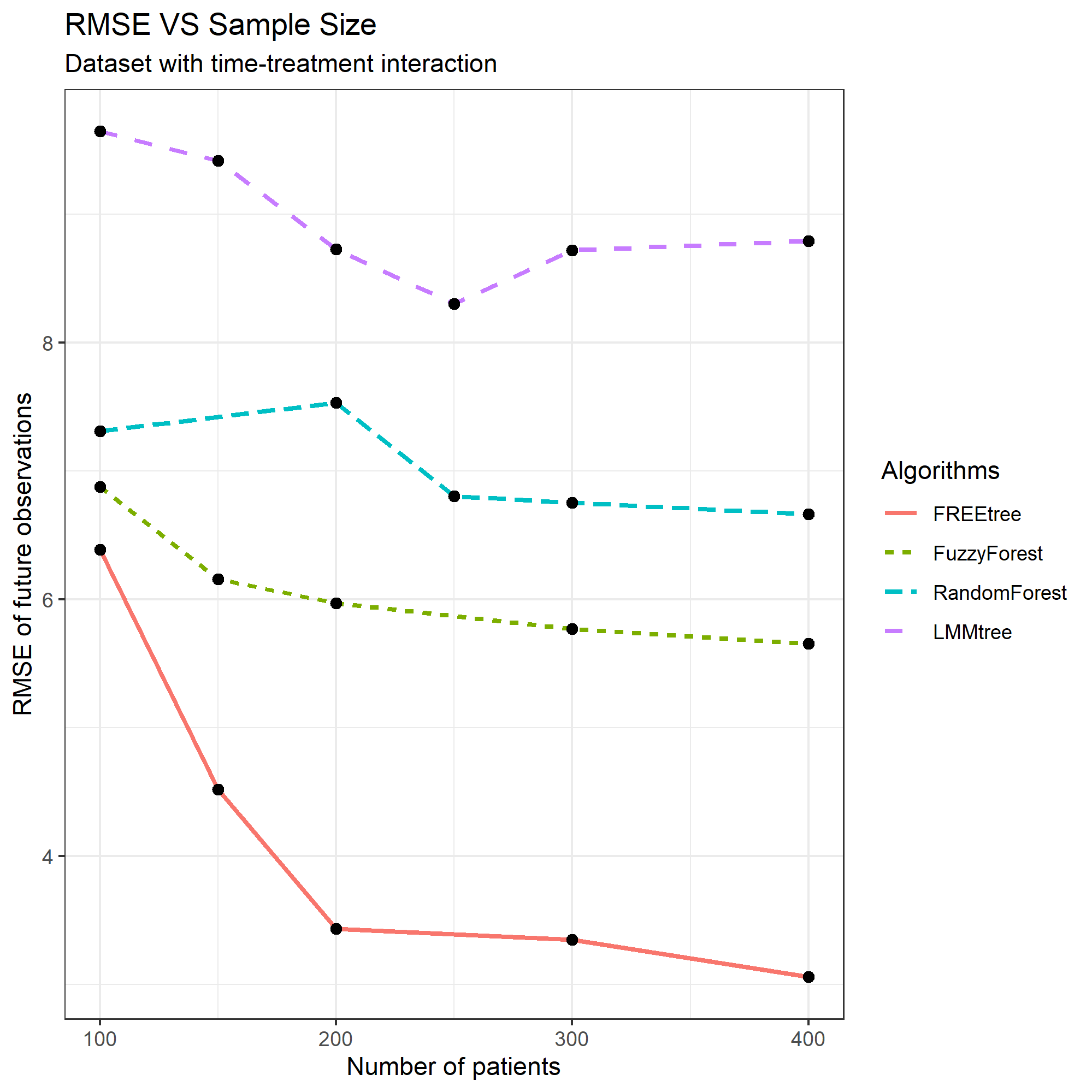}
    \caption{Predictive performance using the time-treatment interaction dataset}
    \label{fig:a1}
\end{figure}

Fig.\ref{fig:a1} shows the results on this dataset. FREEtree outperforms other methods when the sample size is relatively large. When the sample size is relatively small, FREEtree does not have an strong advantage since it has a linear regression model at each leaf, and thus there are many more parameters to estimate, necessitating a larger sample size.

\begin{figure}[htp]
    \centering
    \includegraphics[width=1\textwidth]{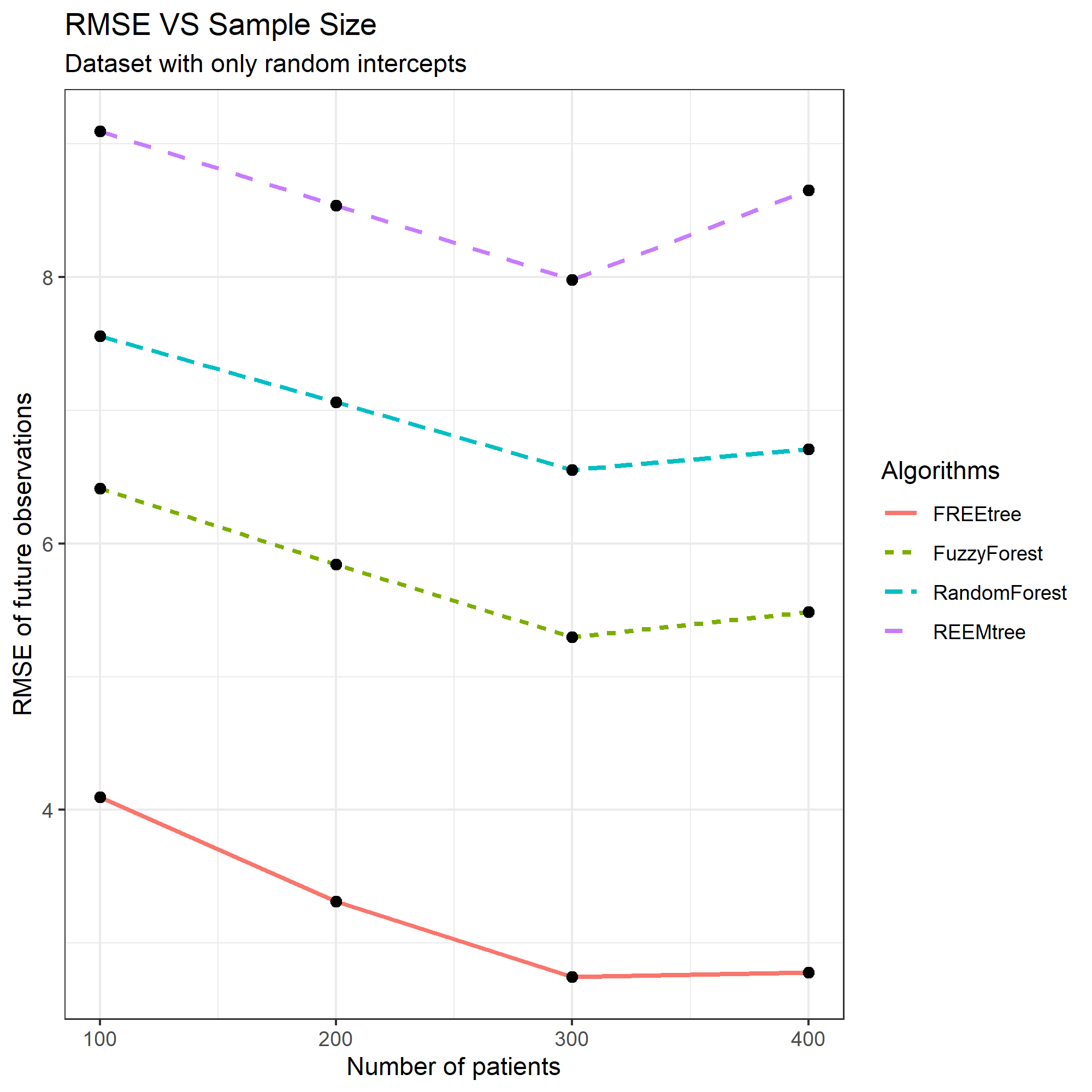}
    \caption{Predictive performance on the dataset with only random intercepts}
    \label{fig:a0}
\end{figure}

Fig.\ref{fig:a0} gives the results of the performance on the simulated dataset with only random intercepts, a special case of longitudinal structure. The RMSE of Random Forests, Fuzzy
Forests, RE-EM tree and FREEtree are given. Only $\{X^{(v)}\}_{v=1}^{400}$ are used in these algorithms. This analysis shows that FREEtree has better predictive performance than other algorithms and performs better when the sample size is larger. Note, that unlike the case in the previous simulation, FREEtree does well even when $n$ is relatively small because the dataset structure here is much simpler.

\subsection{Feature selection performance}
In this section, we compare the performance of feature selection from FREEtree and Fuzzy Forests, which is designed for feature selection. For Fuzzy Forests, we computed the proportion of times each feature was selected as important over 50 simulation runs on the same training set with different seeds and/or tuning parameters. In each run, the top 12 features are selected in the first simulation with time-treatment interaction dataset and top 10 features are chosen in the second simulation using dataset with only random intercepts. For FREEtree, the final chosen features are presented. 

\begin{figure}[htp]
    \centering
    \includegraphics[width=1\textwidth]{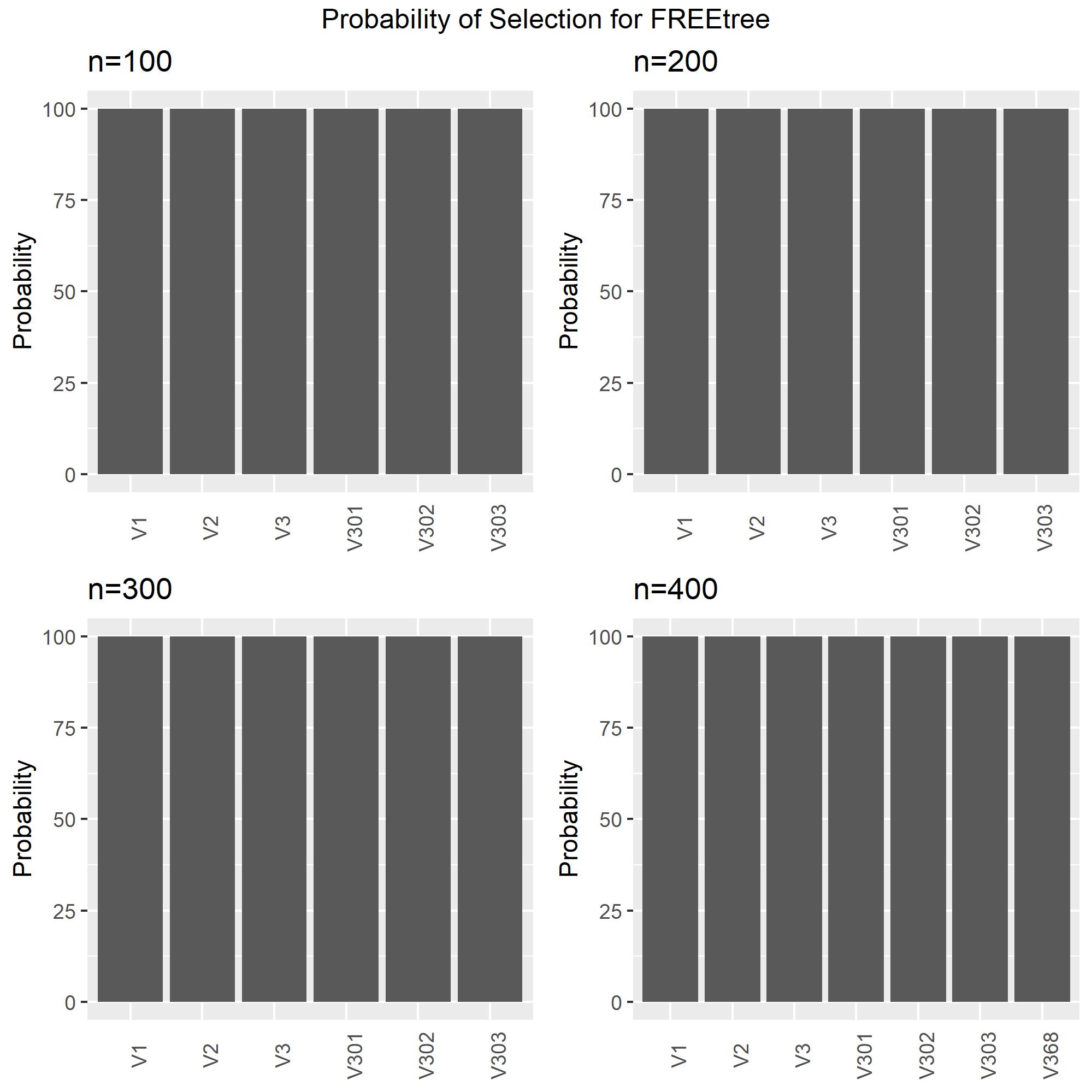}
    \caption{The selected feature of FREEtree with different sample size $n$ on the dataset with time-treatment interaction. }
    \label{fig:a1FREEtree}
\end{figure}

\begin{figure}[htp]
    \centering
    \includegraphics[width=1\textwidth]{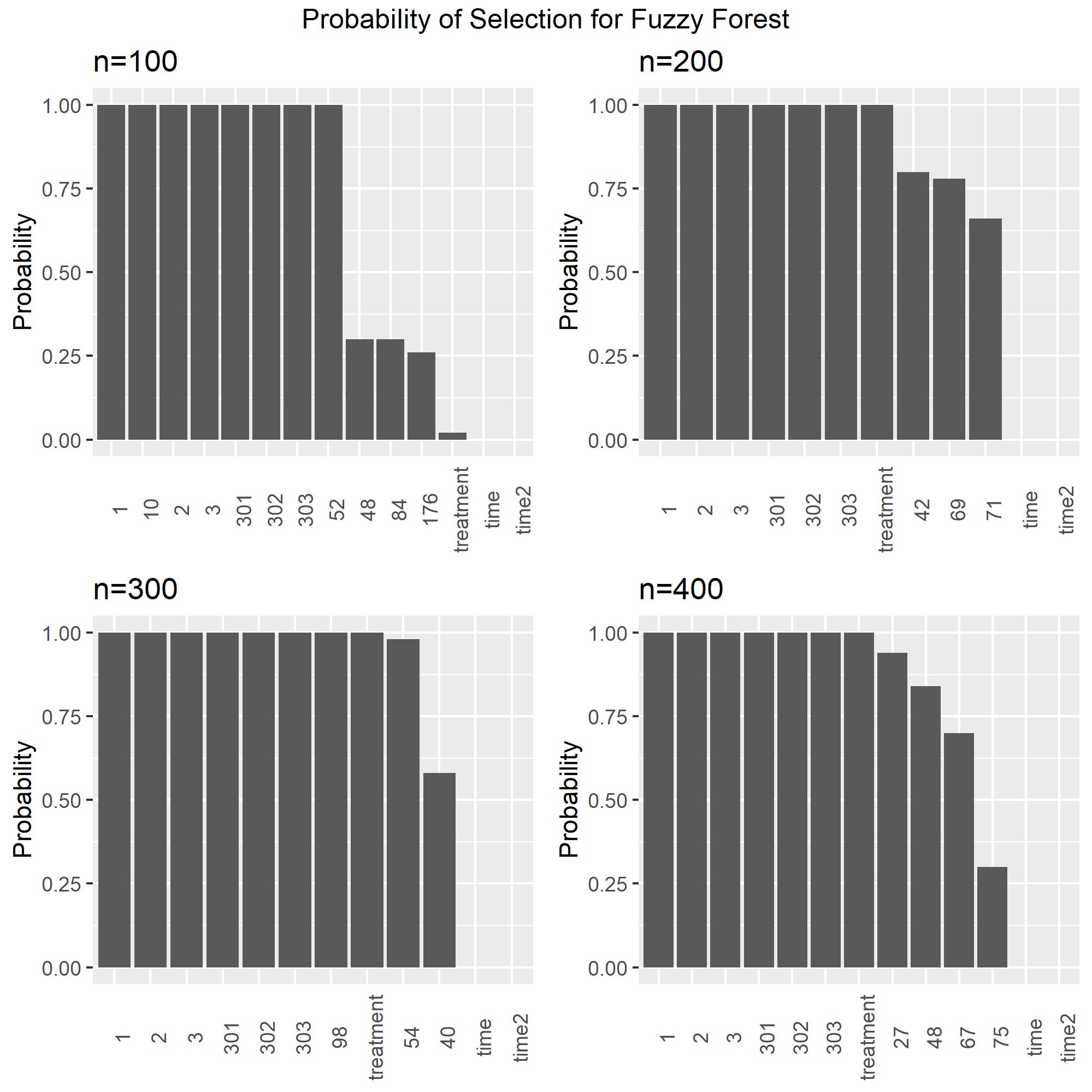}
    \caption{Feature selection performance of Fuzzy Forests on the dataset with a time-treatment interaction}
    \label{fig:a1Fuzzy}
\end{figure}

In the first simulation, shown in Fig.\ref{fig:a1Fuzzy}, where the true features are $X^{(1)},X^{(2)},X^{(3)},X^{(301)},X^{(302)}$\\$,X^{(303)}$, $treatment$, $time$ and $time^2$, Fuzzy Forests successfully identified $X^{(1)},X^{(2)},$\\$X^{(3)},X^{(301)},X^{(302)},X^{(303)}$ with probability 1 but missed $time$ and $time^2$ completely (selected 0 times) regardless of the overall sample size. Fuzzy Forests identifies $treatment$ with probability 1 when $n \geq 150$.  As for FREEtree, since treatment, time and time2 are explicitly specified to use as splitting and regression variable respectively, we only need to examine the final selected features from  \texttt{var\char`_select} $\{X^{(v)}\}_{v=1}^{400}$. Fig.\ref{fig:a1FREEtree} gives results for this simulation and it shows that in this dataset FREEtree can recover the true important features when $n \geq 150$.

\begin{figure}[htp]
    \centering
    \includegraphics[width=1\textwidth]{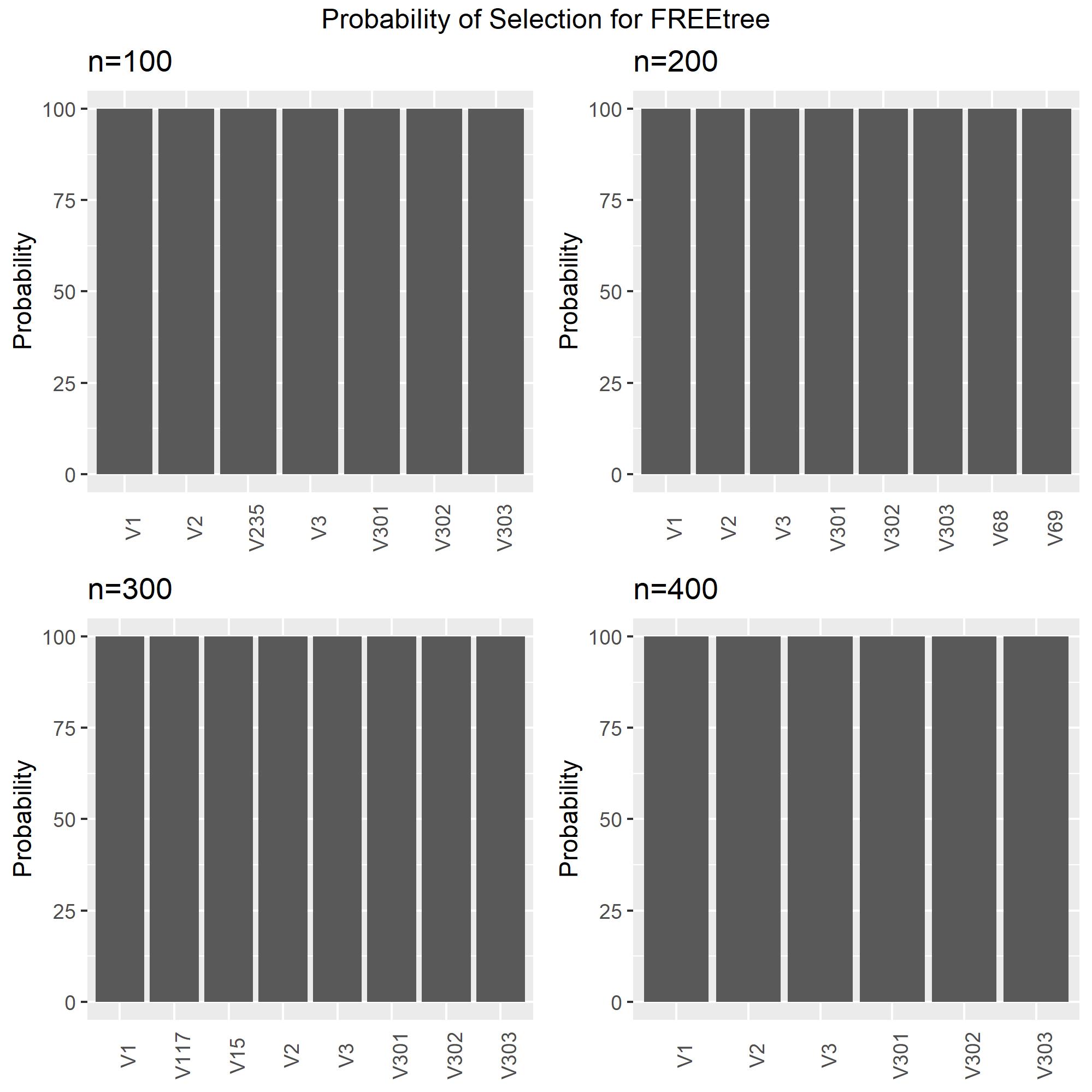}
   \caption{The selected feature of FREEtree with different sample size $n$ on the dataset with only random intercepts. The first column is the number of patients or sample size $n$. }
    \label{fig:a0FREEtree}
\end{figure}

\begin{figure}[htp]
    \centering
    \includegraphics[width=1\textwidth]{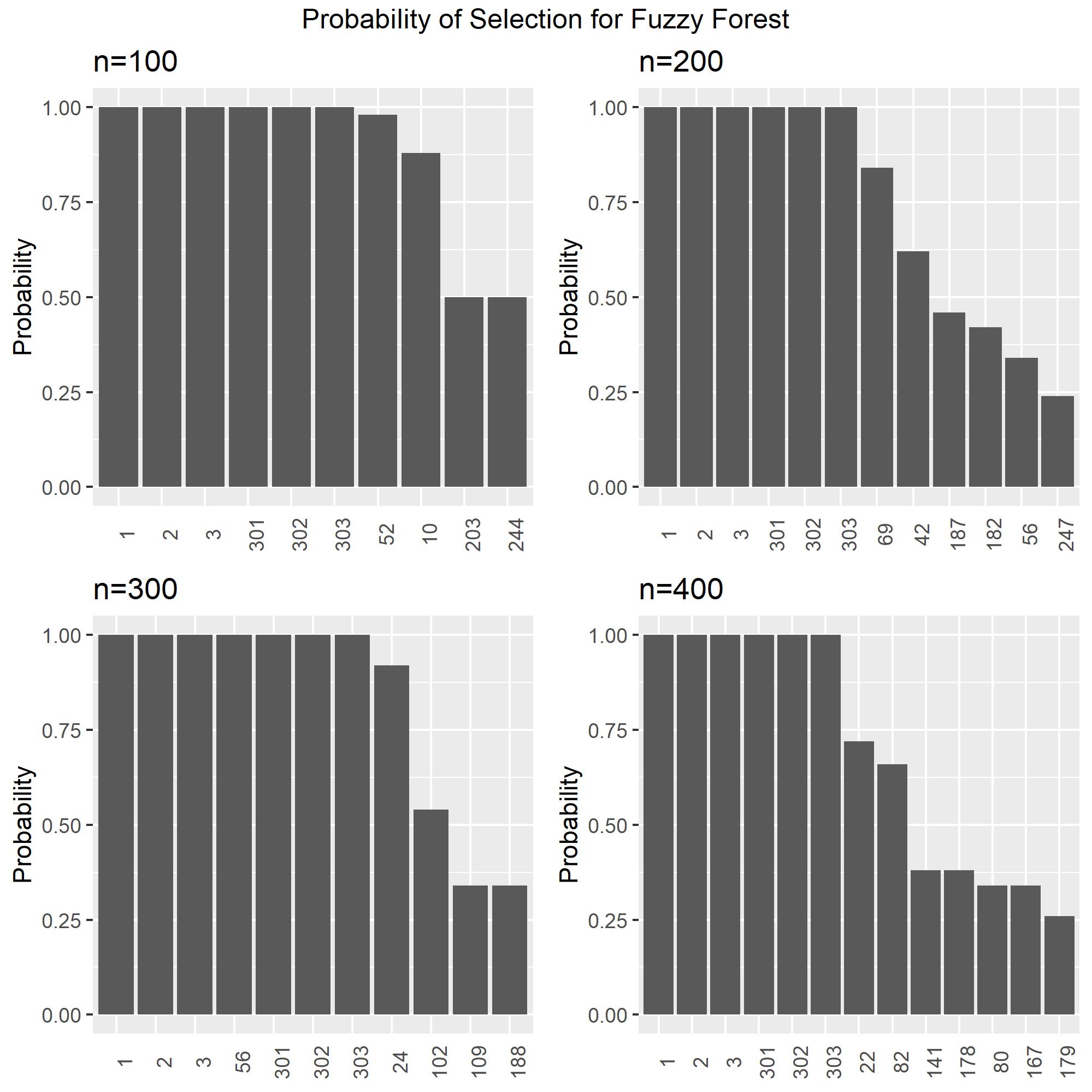}
    \caption{Feature selection performance of Fuzzy Forests on the dataset with only random intercepts}
    \label{fig:a0Fuzzy}
\end{figure}

In the second simulation where the true generating process only includes random intercepts, the feature selection performance of Fuzzy Forests and FREEtree were also studied. Fig.\ref{fig:a0Fuzzy} shows the results of Fuzzy Forests, which recovers all the important features correctly. Fig.\ref{fig:a0FREEtree} shows that FREEtree can also recover all the important features for all of the sample sizes tested. 

\subsection{Estimation of the underlying pattern}
The advantage of FREEtree is not only in its in higher prediction accuracy, but also in how it fits the underlying structure due to the models at its leaves. Recall that in the first simulation, the dataset has a time-treatment interaction. That is, the treatment-time components will first drop then increase for treatment1 and will first increase and then drop for treatment2. In this section we will examine whether FREEtree can recover the true time pattern for different treatments. The underlying true pattern should have the following form:
\[ \begin{cases} 
      (t-3)^2 & \text{treatment}=1 \\
      -(t-3)^2 & \text{treatment}=2
   \end{cases}
\]

FREEtree was able to successfully detect the time-treatment interaction in this simulation. Table \ref{table: mean of time pattern } shows that FREEtree gives a reasonable estimation of the time pattern function. However, note that patterns like this cannot be directly observed using tree-based methods such as RE-EM tree because the leaves in RE-EM tree correspond to an averaged value instead of a model. 

\begin{table}[htp]
\centering
  \begin{tabular}{lSSSSSS}
    \toprule
    \multirow{2}{*}{Sample Size} &
      \multicolumn{2}{c}{treatment1} &
      \multicolumn{2}{c}{treatment2}  \\
      & {time} & {time2} & {time} & {time2}\\ 
      \midrule
    100 & -8.88 & 1.36 & 5.23 & -0.89  \\
    200 & -5.60 & 0.88 & 5.46 & -0.91  \\
    300 & -6.06 & 0.99 & 5.43 & -0.91  \\
    400 & -6.40 & 1.07 & 6.16 & -1.01 \\
    \bottomrule
  \end{tabular}
  \caption{The mean of coefficients of linear models at leaves for each treatment. The coefficients of time and time2 should be 6 and 1 for treatment1 and -6 and -1 for treatment2. }
  \label{table: mean of time pattern }
\end{table}

\section{Application}
We illustrate a real data application of FREEtree in a wide longitudinal dataset of World Bank, IMF and Penn World Table country level economic and developmental indicators. Using the adoption of inflation targeting by a nation's central bank as a  treatment variable, we wish to predict the percentage change in a country's consumer price index (CPI) as a measure of the inflation rate. Merging together 15 different data sources\footnote{IMF World Economic Outlook (October 2019), IMF Financial Development Index Database, Penn World Table version 9.1, and the following World Bank databases: World Development Indicators, Education Statistics, Doing Business, Health Nutrition and Population Statistics, Gender Statistics, Global Financial Development, Health Equity and Financial Protection Indicators, Worldwide Governance Indicators, Worldwide Bureaucracy Indicators, Statistical Capacity Indicators, Global Jobs Indicators and Environment, Social and Governance Data.}, we obtain a final data set of 120 countries with 393 features observed for a 12 year period between 2005 and 2016 inclusively. The data series mostly comprise of population ratios, per capita metrics, year-over-year rates of change, proportions of national accounts, and scaled indicators, before being normalized to have mean zero and unit standard deviations.\\

Country level indicators are often highly correlated across time, with many series being very related to or subsets of others. Although tree-based techniques like Random Forests and Fuzzy Forests can process large numbers of series through feature selection, they do not have the capabilities to model mixed effects or give a single interpretable tree. \pkg{glmertree} can manage such effects while directly incorporating treatment variables into the analysis using GLM at each final node, it cannot handle the number of features in the dataset given the dimensionality problems inherit in linear regression. We compare results obtained by FREEtree to Random Forests and Fuzzy Forests in an example that takes advantage of individual country-level effects, as well as the central bank price targeting policy   \textit{country\_id} was declared to be the subgroup \pkg{cluster}, while \pkg{fixed regress} included a linear and quadratic temporal term. Inflation targeting adoption, a binary variable,  was declared for \pkg{consider split}, the rest of the features were included for the screening and selecting process. The formula takes the form:  
$$  CPI_{i,t} = year_t + year^2_t + treatment_{i,t} + X_{i,t} \;|\; country\_id_i \;|\;  X_{i,t} $$
where $X_{i,t}$ include all the other features to be screened and selected by FREEtree's algorithm.  \\

\begin{figure}[htp]
    \centering
    \includegraphics[width=1.2\textwidth]{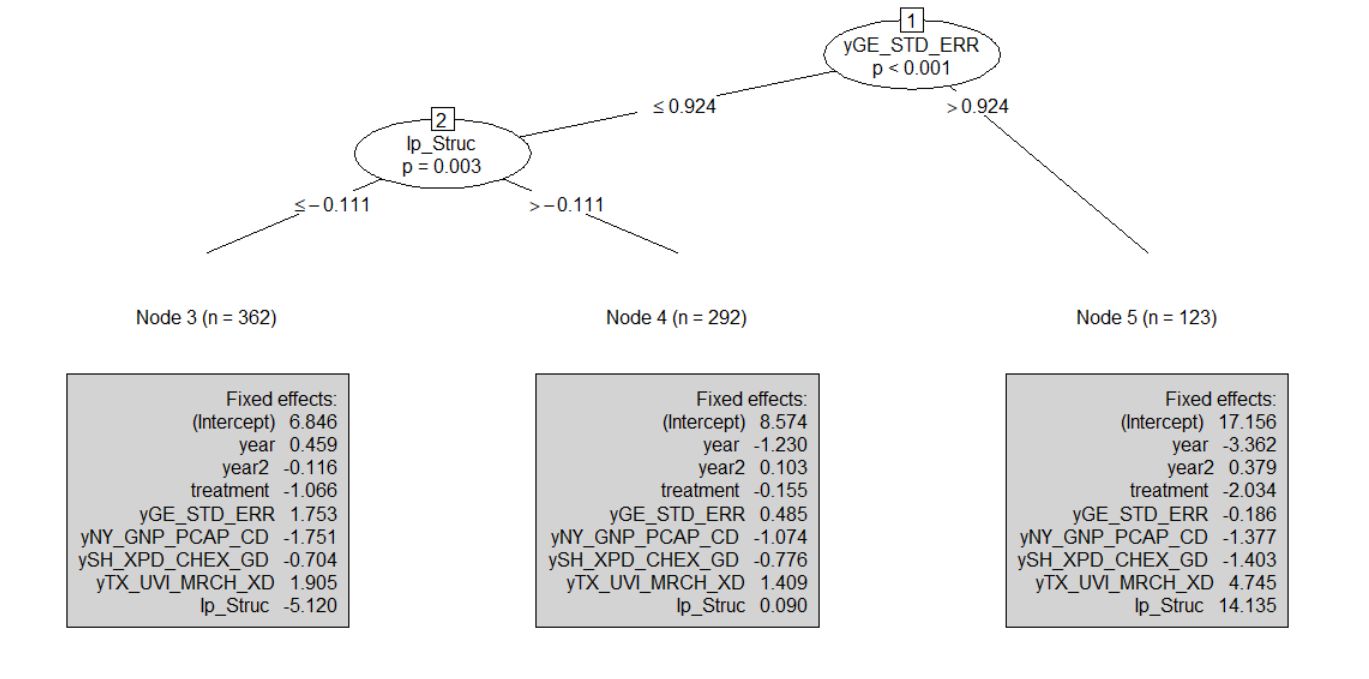}
    \caption{FREEtree model tree applied to real sample data}
    \label{fig:FTresult}
\end{figure}

\begin{figure}[htp]
    \centering
    \includegraphics[width=1.2\textwidth]{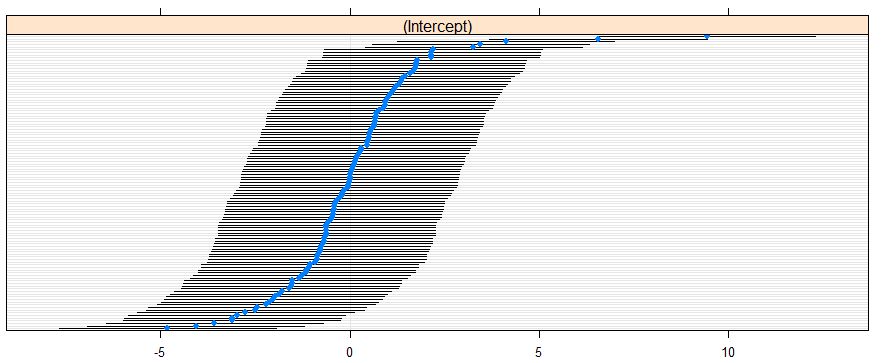}
    \caption{FREEtree model tree applied to real sample data: individual country effects}
    \label{fig:FTresult intercept}
\end{figure}

The resulting tree has three nodes from two split variables (investment price index and GDP volatility) and 9 explanatory variables, including GNP per capital, fuel and GDP volatility (Figure \ref{fig:FTresult}). The mixed effect paint a picture of volatile frontier economies in various states of high inflation or deflation, while industrialized nations tend to be closer to the mean (Figure \ref{fig:FTresult intercept}). \\ 

\begin{figure}[htp]
    \centering
    \includegraphics[width=1.2\textwidth]{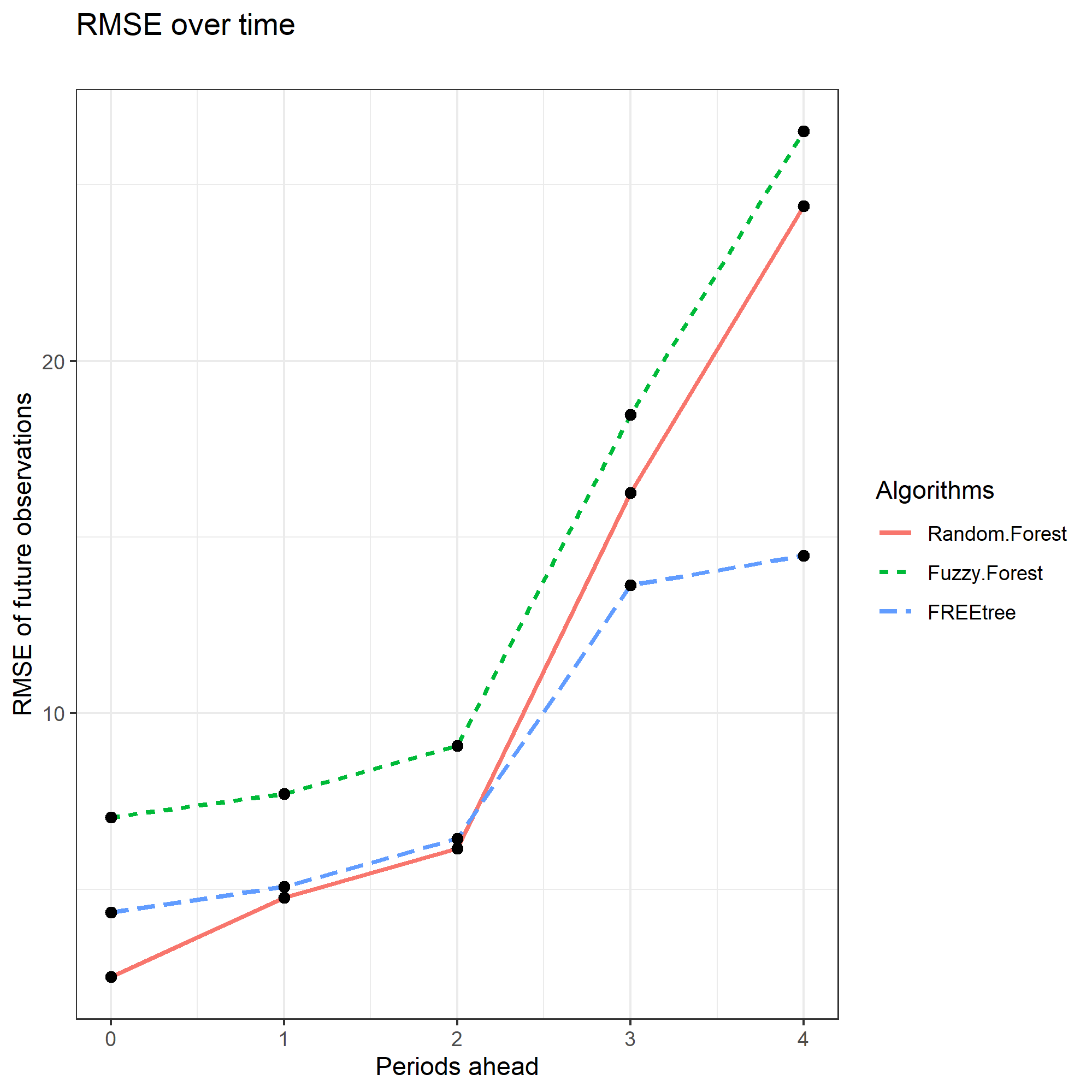}
    \caption{FREEtree model tree applied to real sample data: cross sectional performance (20 test countries)}
    \label{fig:comp_xsectional}
\end{figure}

\begin{figure}[htp]
    \centering
    \includegraphics[width=1.2\textwidth]{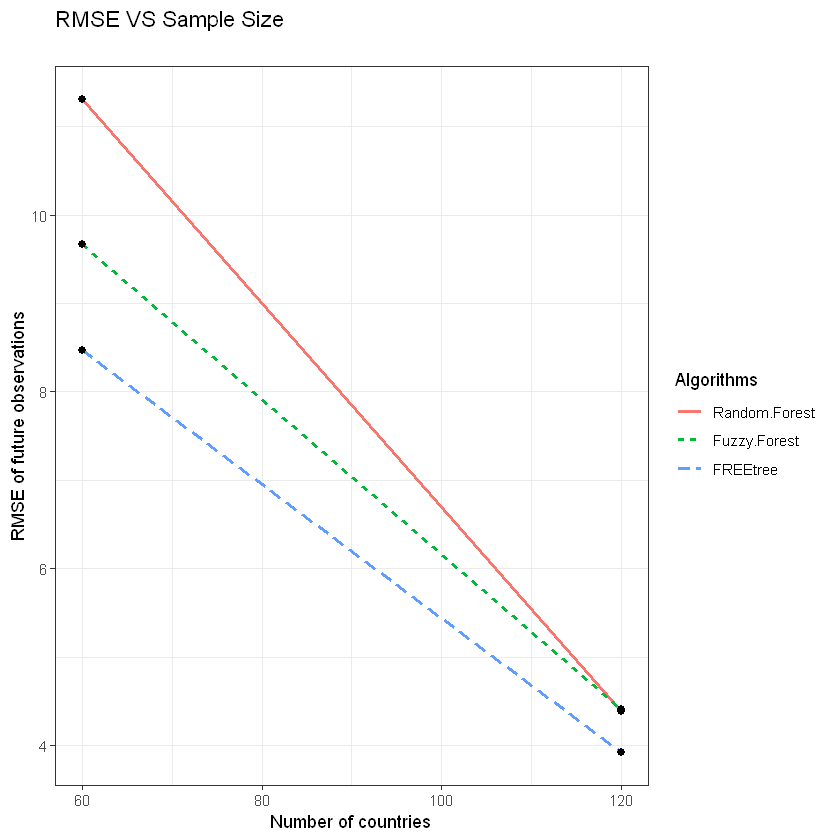}
    \caption{FREEtree model tree applied to real sample data: forward performance by horizon}
    \label{fig:comp_temporal}
\end{figure}

Using mainly default parameter values, WGCNA yielded four modules with 150, 125, 80 and 38 features, and the grey module being the 3rd largest in size. We can see that FREEtree performs notably better in larger samples (Figure \ref{fig:comp_xsectional}) and further out of sample temporally (Figure \ref{fig:comp_temporal}).

\section{Interpretability}
\label{Interpretability}
FREEtree differentiates itself from other model trees in its ability to accept a very large number of features, addressing dimensionality issues when $p >> n$. Although this is a feature it shares in common with Random Forest and Fuzzy Forest, it distinguishes itself from these ensemble methods by being able to produce a single tree the user can readily interpret and understand while also providing superior predictions. \\

The production of a single decision tree also lets the user specify persistent features that will make it into the regression nodes, inherited from LMM tree. In addition, the user can specify subgroup cluster indicators and which features are guaranteed to make it past the screening process as regressors in the linear model. This flexibility caters well to researchers seeking to understand the impact of their variables of interest among a high number of other features, allowing them to effectively customize the output tree while taking advantage of WGCNA-based feature selection.\\ 



\begin{figure}[htp]
    \centering
    \includegraphics[width=1.2\textwidth]{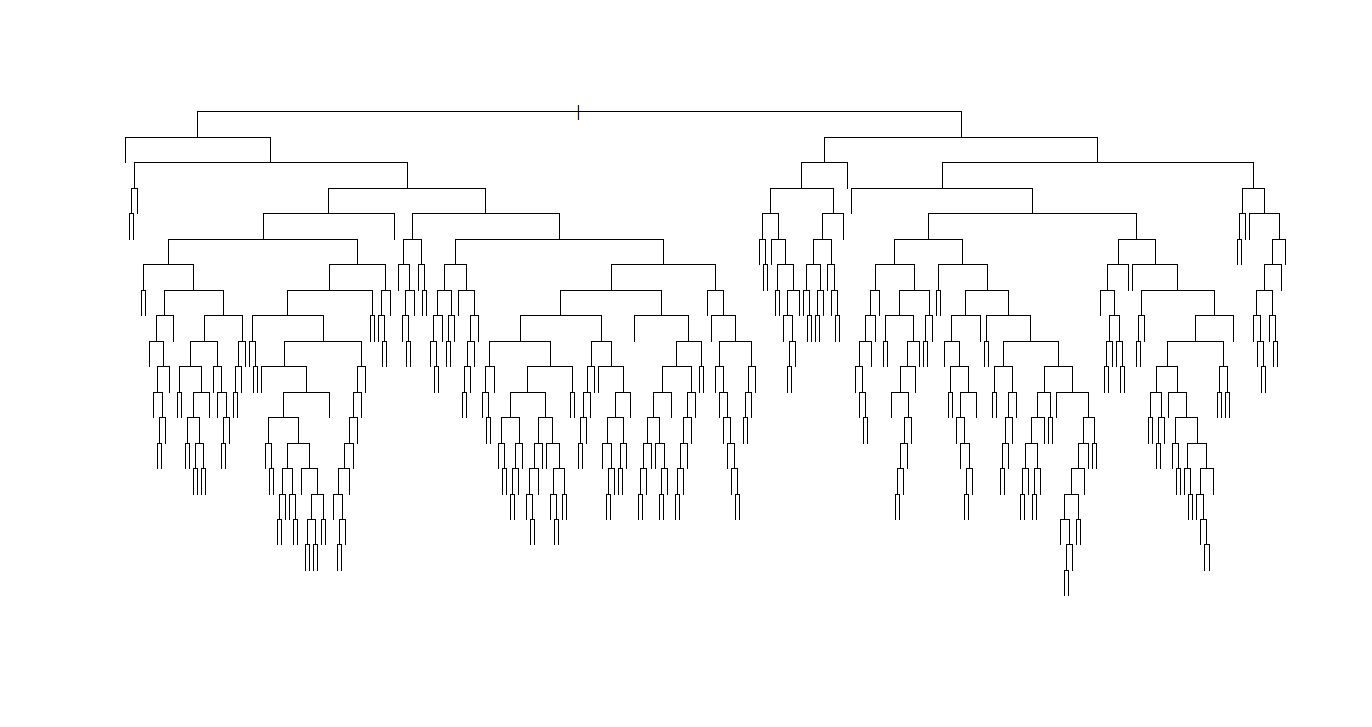}
    \caption{Representative tree of Random Forest applied to real sample data}
    \label{fig:exRF}
\end{figure}

\section{Discussion and Future Research}
\label{Discussion and future work}
At the feature clustering step, FREEtree uses Pearson correlation as the similarity function, which may not be optimal when the measurements of each feature of any patients are time series. That is, for patient $i$ and feature $v$, $X^{(v)}_{i1},X^{(v)}_{i2},..,X^{(v)}_{iT}$ is a time series. In order to cluster features in this case, we have to cluster time series. According to our simulations, where Auto-Regressive and Compound-Symmetric structure were imposed on each feature $X^{(v)}$, WGCNA still works when the correlation between features are relatively large. However, when the correlation is relatively low, WGCNA may not find strong associations and assign all the features to the grey group. One way to get around this is that when doing WGCNA analysis, instead of using correlation of features when building the similarity matrix, we use time series distance measures such as Dynamic time warping (DTW) and average them with respect to each patient and finally transform it into a similarity measure. In this case, the adapted WGCNA can detect module distinctions even if the correlation between features is relatively low. However, it is most be pointed out that computing time series distance measure such as DTW requires a lot of computational resources and in applications where $p$ is really large, replacing correlation with a time series distance measure may not be practical computationally.  

\section{Conclusions}
In this paper we have presented Fuzzy  Random  Effect  Estimation  tree  (FREEtree) algorithm that can provide a relatively unbiased way to do feature selection in the presence of correlation between features. Also, it deals with longitudinal data by using a random effect model tree, where the fixed effect is modelled as a piece-wise linear model, which has greater fitting and predicting power than RE-EM tree.  It is expected that FREEtree can be widely used in application where the data has longitudinal structure as well as many correlated features.

\bibliographystyle{plain}
\bibliography{main}
\end{document}